\PassOptionsToPackage{table}{xcolor}
\documentclass[letterpaper]{article} 
\usepackage{aaai2027}
\nocopyright
\usepackage[hyphens]{url}  
\usepackage{graphicx} 
\usepackage{natbib}  
\usepackage{caption} 
\usepackage{algorithm}
\usepackage{algorithmic}
\usepackage{booktabs}
\usepackage{amsmath}
\usepackage{amssymb}
\usepackage{newfloat}
\usepackage{listings}
\usepackage{tabularx}
\usepackage{comment}
\usepackage{diagbox}
\usepackage{dsfont}
\usepackage{array}
\usepackage{pgf}
\usepackage{colortbl}

\usepackage{algorithm}
\usepackage{algorithmic}
\usepackage{booktabs}
\usepackage{amsmath}
\usepackage{amssymb}
\usepackage{newfloat}
\usepackage{listings}
\usepackage{tabularx}
\usepackage{comment}
\usepackage{diagbox}
\usepackage{dsfont}
\usepackage{array}
\usepackage{pgf}
\usepackage{colortbl}

\usepackage[most]{tcolorbox}

\definecolor{TableBlue}{HTML}{2878B5} 
\definecolor{HeaderBlue}{HTML}{DCECF7}
\usepackage{multirow}       
\usepackage{array}          
\usepackage[table]{xcolor}  
\DeclareCaptionStyle{ruled}{labelfont=normalfont,labelsep=colon,strut=off} 
\floatstyle{ruled}
\newfloat{listing}{tb}{lst}{}
\floatname{listing}{Listing}

\usepackage[table]{xcolor}
\definecolor{gaincolor}{RGB}{30,80,160}
\newcommand{\gain}[1]{{\scriptsize\textcolor{gaincolor}{$_{+#1}$}}}

\definecolor{dropcolor}{RGB}{170,40,40}

\definecolor{TableBlue}{HTML}{2878B5}
\definecolor{HeaderBlue}{HTML}{DCECF7}
\usepackage{multirow}       
\usepackage{array}          
\usepackage[table]{xcolor}  
\DeclareCaptionStyle{ruled}{labelfont=normalfont,labelsep=colon,strut=off} 
\floatstyle{ruled}
\newfloat{listing}{tb}{lst}{}
\floatname{listing}{Listing}

\newcommand{\heatcell}[1]{%
  \ifdim #1pt<2.10pt
    \cellcolor{TableBlue!10}#1%
  \else\ifdim #1pt<2.40pt
    \cellcolor{TableBlue!18}#1%
  \else\ifdim #1pt<2.70pt
    \cellcolor{TableBlue!27}#1%
  \else\ifdim #1pt<3.00pt
    \cellcolor{TableBlue!36}#1%
  \else\ifdim #1pt<3.30pt
    \cellcolor{TableBlue!45}#1%
  \else\ifdim #1pt<3.60pt
    \cellcolor{TableBlue!55}#1%
  \else\ifdim #1pt<3.80pt
    \cellcolor{TableBlue!64}#1%
  \else
    \cellcolor{TableBlue!72}#1%
  \fi\fi\fi\fi\fi\fi\fi
}

\newcommand{\acccell}[1]{%
  \ifdim #1pt<25pt
    \cellcolor{TableBlue!8}#1%
  \else\ifdim #1pt<30pt
    \cellcolor{TableBlue!16}#1%
  \else\ifdim #1pt<35pt
    \cellcolor{TableBlue!24}#1%
  \else\ifdim #1pt<40pt
    \cellcolor{TableBlue!32}#1%
  \else\ifdim #1pt<45pt
    \cellcolor{TableBlue!40}#1%
  \else\ifdim #1pt<50pt
    \cellcolor{TableBlue!48}#1%
  \else\ifdim #1pt<55pt
    \cellcolor{TableBlue!56}#1%
  \else\ifdim #1pt<60pt
    \cellcolor{TableBlue!64}#1%
  \else
    \cellcolor{TableBlue!72}#1%
  \fi\fi\fi\fi\fi\fi\fi\fi
}

\newcommand{\impcell}[1]{%
  \ifdim #1pt<25pt
    \cellcolor{TableBlue!8}#1%
  \else\ifdim #1pt<35pt
    \cellcolor{TableBlue!16}#1%
  \else\ifdim #1pt<45pt
    \cellcolor{TableBlue!26}#1%
  \else\ifdim #1pt<55pt
    \cellcolor{TableBlue!37}#1%
  \else\ifdim #1pt<60pt
    \cellcolor{TableBlue!47}#1%
  \else\ifdim #1pt<65pt
    \cellcolor{TableBlue!57}#1%
  \else\ifdim #1pt<70pt
    \cellcolor{TableBlue!66}#1%
  \else
    \cellcolor{TableBlue!75}#1%
  \fi\fi\fi\fi\fi\fi\fi
}

\usepackage{booktabs}

\title{ArtECulture: Benchmarking Culture-Conditioned Visual Emotion Understanding in Multimodal Large Language Models}

\author{
    Xiaolin Chen\textsuperscript{\rm 1},
    Xuemeng Song\textsuperscript{\rm 2},
    Wenhao Shi\textsuperscript{\rm 1},
    Xianjing Han\textsuperscript{\rm 3},
    Mong-Li Lee\textsuperscript{\rm 1},
    Wynne Hsu\textsuperscript{\rm 1}
}
\affiliations{
    \textsuperscript{\rm 1}National University of Singapore \\
    \textsuperscript{\rm 2}Southern University of Science and Technology\\
    \textsuperscript{\rm 3}Nanyang Technological University\\
}

\begin{document}

\maketitle

\newtcolorbox{promptbox}[2][]{
  enhanced, breakable,
  colback=PromptBackground,
  colframe=PromptRule,
  colbacktitle=PromptBlue,
  coltitle=white,
  fonttitle=\bfseries\footnotesize,
  fontupper=\footnotesize,
  title={#2},
  boxrule=0.6pt, arc=2mm,
  left=3mm, right=3mm, top=2.5mm, bottom=2.5mm,
  toptitle=1mm, bottomtitle=1mm,
  before skip=10pt, after skip=12pt,   
  parbox=false,                         
  #1
}

\newcommand{\placeholder}[1]{\texttt{\textless #1\textgreater}}

\begin{abstract}

Existing visual emotion understanding methods typically ignore cultural variations in emotional perception. We introduce culture-conditioned visual emotion understanding, a task that predicts the culture-specific emotional perception of a given image and explains the underlying rationale.
Although related benchmarks exist, they are limited by inconsistent individual annotations, which hinder the derivation of majority-supported culture-level emotion labels, and imbalanced cultural coverage. Thus, we present ArtECulture, a benchmark containing 6,792 artworks with culture-specific emotion labels and explanations across English, Chinese, and Arabic cultures, with balanced Western and non-Western content. 
Evaluations of 16 open- and closed-source Multimodal Large Language Models (MLLMs) under a zero-shot setting reveal that the task remains challenging, with the best model achieving below 50\% accuracy.  To address this limitation, we introduce a retrieval-augmented culture-conditioned emotion understanding framework, which leverages a concept-based cultural emotion knowledge base to inject explicit cultural knowledge into MLLMs without additional training. The framework improves both culturally aligned emotion prediction and grounded explanation generation. Our benchmark and code will be publicly released.
\end{abstract}

\section{Introduction}
\begin{figure}[!t]
    \centering
    \includegraphics[scale=0.42]{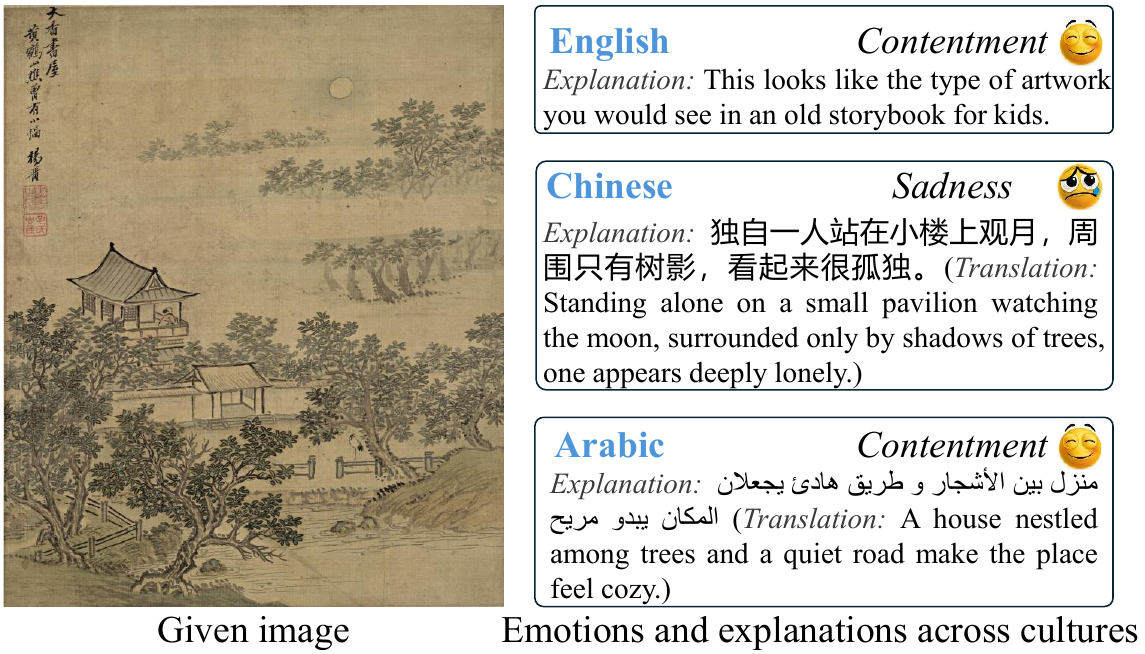}
    \caption{Example from ArtECulture. Viewers from different cultures perceive different emotions from the same image and provide corresponding culture-specific explanations.}
    \label{Model_task}
\end{figure}

{As Multimodal Large Language Models (MLLMs) become  integrated into conversational assistants, human-AI interactions are evolving from text-centric  toward multimodal exchanges, where assistants can retrieve and generate images in their responses. Images, as a rich medium for human communication, can evoke diverse emotional perceptions across users, shaping their experiences and interactions with AI systems. 
Thus, enabling MLLMs to understand such perceptions is essential for building more empathetic and user-aligned assistants.}
However, existing visual emotion understanding methods~\cite{DBLP:conf/iciap/AslanCDMSV22,DBLP:conf/iccv/YangHDLC023,DBLP:conf/naacl/BhattacharyyaW25} assume that images evoke universal emotions across viewers, overlooking the cultural factors that shape emotional perception~\cite{3b6a0c41-318f-3d9d-beed-23b09ba4f207}. As shown in Figure~\ref{Model_task}, the same image may evoke \textit{sadness} among Chinese viewers, who interpret the solitary moon-watching person as loneliness, but \textit{contentment} among English and Arabic viewers.

\begin{figure*}[!t]
    \centering
    \includegraphics[scale=0.7]{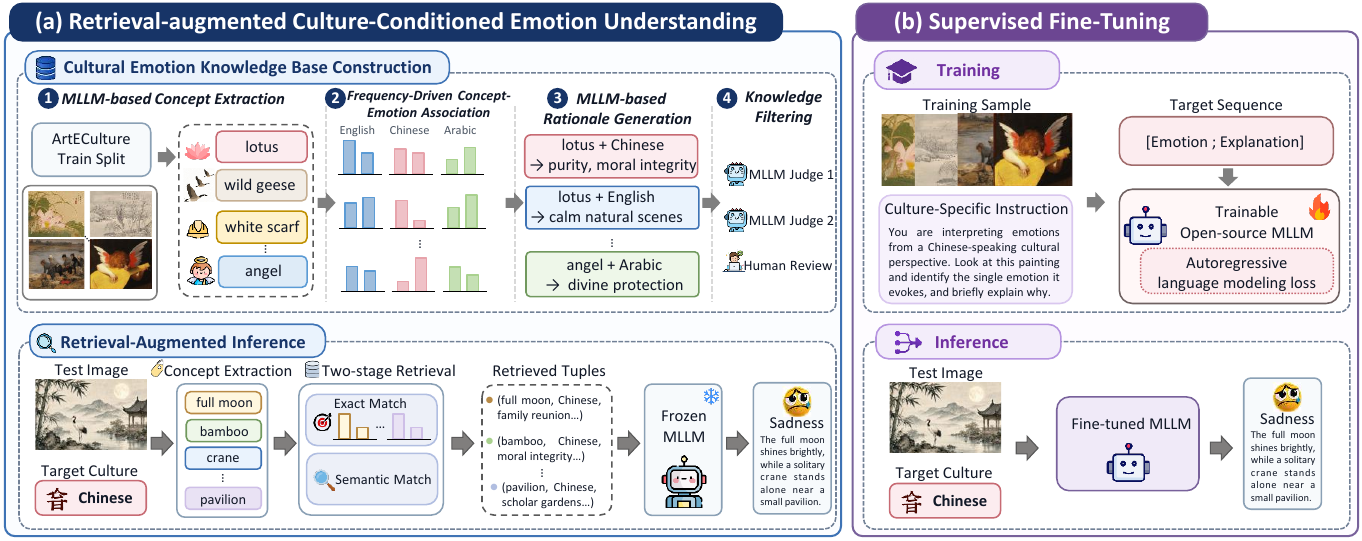}
    \caption{Proposed paradigms: \textit{retrieval-augmented culture-conditioned emotion understanding}  and \textit{supervised fine-tuning}.}
    \label{Model}
\end{figure*}

Recent studies have started to incorporate cultural factors into emotion understanding. Early works introduce benchmarks, e.g., ArtELingo~\cite{DBLP:conf/emnlp/MohamedAAL00E22} and ArtELingo-28~\cite{DBLP:conf/emnlp/MohamedLAH00E24}, which provide culturally diverse emotional interpretations of images across languages and cultures, with each image paired with an emotion label and an explanatory caption. However, these tasks assume that the target emotion is given and focus on emotion-conditioned caption generation rather than emotion prediction. More recent works, including CuLEmo~\cite{DBLP:conf/acl/BelayAGASKY25} and CEDAR~\cite{DBLP:conf/acl/DaiSGLJWLGH26}, investigate culture-conditioned emotion prediction. Nevertheless, they mainly study textual contexts rather than visual perception: CuLEmo is purely text-based, while CEDAR's limited multimodal subset ($\sim400$ images)  uses images as illustrations of textual narratives, capturing emotions inferred from descriptions rather than those perceived from visual content. Thus, we define the task of \textbf{culture-conditioned visual emotion understanding}, which requires predicting the emotion perceived by viewers from a given culture when observing an image and generating the corresponding cultural rationale.

Evaluating this task requires a benchmark with two key properties. First, each image should have culture-level emotion labels that capture the dominant emotional perception within each culture, enabling evaluation of culturally shared patterns and cross-cultural differences. Second, the benchmark should maintain balanced cultural coverage of visual content to avoid evaluation bias toward dominant cultures. ArtELingo is the closest existing resource, providing large-scale culturally diverse image-emotion-caption pairs that can potentially support our task. However, it does not satisfy these requirements. Its individual-level annotations do not yield culture-level labels, as only 40.22\% of Arabic images and 43.13\% of English images have a majority-perceived emotion. Moreover, 79.75\% of its images depict Western art, resulting in imbalanced cultural coverage.

We therefore construct ArtECulture, the first benchmark satisfying both requirements, by reorganizing ArtELingo and adding culturally diverse artworks with new annotations across English, Chinese, and Arabic cultures. Specifically, we retain ArtELingo samples with dominant emotional perceptions within each culture and collect additional non-Western artworks annotated by native speakers to achieve balanced cultural coverage. Finally, ArtECulture contains 6,792 artworks and 92,062 culture-specific emotion-explanation annotations across English, Chinese, and Arabic cultures, with a balanced distribution of Western (55.89\%) and non-Western (44.11\%) content. 

Using ArtECulture, we evaluate $16$ open- and closed-source MLLMs and obtain two findings. 
First, the task remains challenging, as the best-performing model achieves an overall accuracy below $50\%$. 
Second, current MLLMs exhibit English-centric affective priors, with every model performing best on English and nearly all performing worst on Arabic. Motivated by these findings and the observation that cultural emotional perception is highly context-dependent and requires culture-specific knowledge, we propose a training-free retrieval-augmented framework that enhances MLLMs with concept-level cultural emotion knowledge extracted from the training set of ArtECulture.
Specifically, we construct a cultural emotion knowledge base, where each entry associates a visual concept with culture-specific emotions and rationales. Given an image and target culture, the framework retrieves relevant concepts and their associated cultural knowledge to guide MLLMs in emotion prediction and explanation generation. 
As a comparison, we also fine-tune open-source MLLMs on ArtECulture.
The two paradigms show a complementary trade-off: fine-tuning achieves the highest prediction accuracy but degrades explanation quality possibly because of the multiple-reference training setup, whereas  retrieval augmentation improves both prediction and explanations while enabling deployment on closed-source models. 

Our contributions are summarized as follows:
\begin{itemize}
    \item We study culture-conditioned visual emotion understanding, a new task that requires predicting the emotion an image evokes in a given culture and explaining why the emotion arises. 
    We present ArtECulture, the first benchmark for this task, in which every image carries an emotion label  for each culture, with a balanced distribution of Western and non-Western content.
    \item We evaluate $16$ MLLMs on ArtECulture and show that the task is far from solved and that every model is strongest on English, revealing pronounced English-centric affective priors in current MLLMs.
    \item We build a cultural emotion knowledge base and propose a training-free retrieval-augmented pipeline that injects it into MLLMs. Comparing it with supervised fine-tuning reveals a complementary trade-off between prediction accuracy and explanation quality.
\end{itemize}

\section{Related Work}

\textbf{Visual emotion understanding} predicts the emotions evoked by visual content.
Existing datasets pair web or artistic images with discrete emotion labels~\cite{DBLP:conf/mm/MachajdikH10,10.1145/2502081.2502282,DBLP:conf/cvpr/PengCSG15,DBLP:conf/aaai/YouLJY16,DBLP:conf/eccv/PandaZLLLR18,DBLP:conf/nips/MertensYBSV24,zhang2025mme},  explanations~\cite{mohammad-kiritchenko-2018-wikiart,mathews2016aaai-senticap,9577962,9878621}, or emotion-related attributes~\cite{3298239.3298274,DBLP:conf/iccv/YangHDLC023}.
Beyond early hand-crafted features~\cite{DBLP:conf/iciap/AslanCDMSV22}, recent work adapts MLLMs via instruction tuning~\cite{Xie2024EmoVITRE,DBLP:journals/corr/abs-2406-16442,Chen2026MultiEmoBenchMV,wu2026mveiemobserverempowering,10.1609/aaai.v40i3.37184} or benchmarks them on evoked emotions~\cite{DBLP:conf/naacl/BhattacharyyaW25} and emotional intelligence~\cite{DBLP:journals/corr/abs-2502-04424}. However, these efforts collapse annotations from different cultural backgrounds into a single label per image, overlooking the cultural dependence of emotional perception.


\noindent\textbf{Culture-aware emotion understanding} incorporates cultural factors into emotion understanding. 
The first line studies affective captioning: ArtELingo~\cite{DBLP:conf/emnlp/MohamedAAL00E22} and ArtELingo-28~\cite{DBLP:conf/emnlp/MohamedLAH00E24} extend ArtEmis with multicultural annotations for affective captioning, where emotions are provided as inputs rather than prediction targets.
The second line predicts emotions under cultural or linguistic conditions from text.
Multilingual resources~\cite{DBLP:conf/acl/MuhammadOAWRBKS25,muhammad-etal-2025-semeval}  cover emotion detection in $28$ languages, but each text is tied to a single language and lacks cross-cultural judgments of the same stimulus.
CuLEmo~\cite{DBLP:conf/acl/BelayAGASKY25} examines cultural perception with text-only inputs, and CEDAR~\cite{DBLP:conf/acl/DaiSGLJWLGH26} uses a limited image subset illustrating textual narratives.
These benchmarks focus on single-label evaluation without explanations and leave model adaptation unexplored. In contrast, we target culture-conditioned visual emotion perception, requiring both emotion prediction and cultural rationales, and investigate how MLLMs can be enhanced for this task.

\noindent\textbf{Cultural understanding of MLLMs} examines whether MLLMs hold knowledge about diverse cultures.
Benchmarks~\cite{DBLP:conf/emnlp/0001BPRCE21,DBLP:conf/nips/RomeroLWGMPOVBJ24,nayak-etal-2024-benchmarking,Vayani2024AllLM,DBLP:conf/acl/0001HBL25,tan-etal-2026-blend,DBLP:conf/aaai/WangLYHLWCC25}  test factual cultural knowledge about food, clothing, rituals, and landmarks, revealing consistent gaps on non-Western cultures, which follow-up studies mitigate via cultural training data~\cite{nyandwi-etal-2025-grounding} or retrieval~\cite{li2026ravenea,NEURIPS2020_6b493230}.
Unlike these works on objective cultural facts, we study subjective emotional perceptions elicited by visual content across cultures, with a knowledge base encoding concept-level emotional conventions rather than factual knowledge.

\section{ArtECulture Benchmark}
\label{sec:benchmark}
We construct ArtECulture in two steps: re-curating ArtELingo into a base pool with majority-supported culture-level labels, and collecting non-Western artworks with new annotations to rebalance the regional distribution.

\begin{table}[t]
\centering
\footnotesize
\setlength{\tabcolsep}{2.7pt}
\renewcommand{\arraystretch}{1.0}
\caption{Detailed statistics of  ArtECulture.}
\label{tab:arteculture-statistics}
\begin{tabularx}{\linewidth}{Xrrrr}
\toprule
Item & Train & Valid & Test & Total \\
\midrule  
\#Images & 5,434 & 679 & 679 & 6,792 \\
\#Annotations across  cultures & 73,785 & 9,082 & 9,195 & 92,062 \\
\#Average annotations per image  & 13.58 & 13.38 & 13.54 & 13.55 \\
\midrule  
\#English annotations & 24,835 & 3,037 & 3,060 & 30,932 \\
\#Chinese annotations & 25,256 & 3,144 & 3,159 & 31,559 \\
\#Arabic annotations & 23,694 & 2,901 & 2,976 & 29,571 \\
\bottomrule
\end{tabularx}
\end{table}



\subsection{Re-curation of ArtELingo}
\label{sec:artelingo}
Each ArtELingo image is annotated by multiple annotators from each culture. We retain an image only when majority agreement is achieved separately within all three cultures, with the majority emotion in each culture assigned as its culture-level label.
This yields $4{,}928$ images with  emotion-explanation annotations  from all three cultures, comprising $3{,}796$ Western ($77.03\%$) and $1{,}132$ non-Western ($22.97\%$) artworks.
However, the pool remains Western-centric, requiring non-Western artwork augmentation.

\subsection{Non-Western Artwork Augmentation}
\label{sec:nonwestern}

\paragraph{Collection.}
We collect additional non-Western artworks from the public benchmark VULCA-BENCH~\cite{yu2026vulcabench}, a multicultural art-critique benchmark  drawn from the open collections of authoritative museums.
In total, we obtain $2{,}783$ artworks spanning five major traditions, namely Chinese, Japanese, Korean, Indian, and Islamic art.

\paragraph{Annotation.}
We recruit annotators from Prolific\footnote{\url{https://www.prolific.com}.} with the requirement that the target language is their first language.
Following ArtELingo~\cite{DBLP:conf/emnlp/MohamedAAL00E22} and CEDAR~\cite{DBLP:conf/acl/DaiSGLJWLGH26}, native language serves as the criterion of cultural membership, and each culture-level label reflects the dominant perception within the corresponding language community.
In each community, around $70\%$ of annotators reside in one dominant country ($69.1\%$ in the UK for English, $74.6\%$ in China for Chinese, and $73.9\%$ in Egypt for Arabic); full demographics are provided in the supplementary material.
As in ArtELingo, each annotator  selects one emotion from nine categories (\emph{i.e.,} \textit{contentment}, \textit{awe}, \textit{amusement}, \textit{sadness}, \textit{fear}, \textit{excitement}, \textit{disgust}, \textit{anger}, and \textit{other}), and writes a short explanation in their native language.
For each image-culture pair, we collect three independent annotations and adopt the majority emotion as the culture-level label; if no majority is reached, two additional annotators are recruited and the vote is taken over the five annotations, discarding images that still lack a majority.
Importantly, annotations are collected independently within each culture to preserve cross-cultural divergence.
This process retains $1{,}864$ of the $2{,}783$ collected images, yielding $17{,}526$ emotion-explanation annotations across the three cultures.

\subsection{Statistics and Insights}
\paragraph{Dataset Statistics.}
Table~\ref{tab:arteculture-statistics} summarizes the statistics of ArtECulture. ArtECulture contains $6{,}792$ artworks, where $4{,}928$ are retained from ArtELingo and $1{,}864$ are newly collected, raising the non-Western proportion from $20.25\%$ in ArtELingo to $44.11\%$.
It provides $92{,}062$ emotion-explanation annotations, averaging about $4.5$ annotations per image per culture, with a roughly balanced distribution across English ($30{,}932$), Chinese ($31{,}559$), and Arabic ($29{,}571$).

\paragraph{Agreement Analysis.}
Within each language community, nominal Krippendorff's $\alpha$ is $0.309$, $0.324$, and $0.306$ for English, Chinese, and Arabic, comparable to ArtELingo~\cite{DBLP:conf/emnlp/MohamedAAL00E22}.
On average, the majority emotion receives $70.81\%$, $76.32\%$, and $70.13\%$ of the votes per image, respectively. 
In contrast, only $924$ images ($13.60\%$) receive the same emotion across all three cultures, while $1{,}381$ ($20.33\%$) exhibit three distinct emotions (Fleiss' $\kappa=0.088$ treating  three culture-level labels as raters).
The contrast between within- and cross-culture agreement confirms  the labels capture consistent perceptions within each community that genuinely  diverge across cultures, motivating  culture-conditioned visual emotion understanding.



\paragraph{Data Split.}
The samples are split into training, validation, and test sets in an 8:1:1 ratio, preserving the overall Western/non-Western distribution.
Since each image carries labels from all three cultures, the same split is applied across cultures.
Detailed statistics and analysis of ArtECulture  are provided in the supplementary material.

\section{Method}


\subsection{Problem Formulation}
Suppose we have a set of images $\mathcal{V}=\{v_1,v_2,\cdots,v_M\}$ and a set of cultural groups $\mathcal{C}=\{c_1,c_2,\cdots,c_N\}$, where $M$ and $N$ are the number of images and cultures, respectively.
Each $c_j$ is  operationalized as a native-speaker language community following prior work~\cite{DBLP:conf/emnlp/MohamedAAL00E22,DBLP:conf/acl/DaiSGLJWLGH26}.
All cultures share a predefined set of discrete emotion categories $\mathcal{E}=\{e_l\}_{l=1}^{L}$, where $L$ is the number of categories.
For each culture $c_j$, every image $v_i$ is associated with a culture-level emotion label $e_i^j\in\mathcal{E}$ and a set of reference explanations $\mathcal{S}_i^j=\{s_{i,k}^{j}\}_{k=1}^{K_i^j}$ that support the assigned emotion label.
Given an image $v_i$ and target culture $c_j$, the goal is to build a model $\mathcal{F}$ that predicts the perceived emotion $\hat{e}_i^j$ and generates its supporting explanation $\hat{s}_i^j$.


\subsection{Retrieval-augmented Culture-Conditioned  Emotion Understanding}
\label{sec:rag}
Existing MLLMs still struggle with culture-conditioned emotion understanding without access to explicit cultural knowledge, as  our experiments will demonstrate, motivating the  use  of external cultural knowledge. 
Inspired by retrieval-augmented generation~\cite{NEURIPS2020_6b493230}, we propose retrieval-augmented culture-conditioned emotion understanding, a training-free framework that augments a frozen MLLM with retrieved cultural emotion knowledge.

\subsubsection{Cultural Emotion Knowledge Base.}
\label{sec:kb}
Since cultural differences in emotional responses are often reflected through key visual concepts (\emph{e.g.,} moon) that carry culture-specific meanings, effective cultural emotion knowledge should be concept-grounded, capturing both associated emotions and their underlying cultural rationales.
Accordingly, we define each knowledge entry as a tuple $(\textit{concept},\,\textit{culture},\,\textit{dominant emotion},\,\textit{rationales})$. 
We construct the knowledge base from the training set of ArtECulture through the following four steps. 

 1) \textit{MLLM-Based Concept Extraction.}
This step collects  visual concepts that may carry cultural meanings.
We prompt an MLLM (Gemini~3.5 Flash) to list  salient concepts of each training image, covering objects, colors, scenes, and cultural motifs, e.g., ``full moon'', and ``crane''. 
Merging concepts of all images gives a concept vocabulary $\mathcal{Q}=\{q_1,\dots,q_R\}$, where $q_r$ is the $r$-th concept and $R$ is the vocabulary size.

2) \textit{Frequency-Driven Concept-Emotion Association.}
This step mines concept-emotion pairs based on co-occurrence frequency statistics. 
Let $\mathcal{N}_r$ denote the set of training images that contain concept $q_r$.
For a culture $c_j$, the probability that $q_r$ evokes an emotion $e\in\mathcal{E}$ is computed as the fraction of images in $\mathcal{N}_r$ whose culture-level label under $c_j$ is $e$,
\begin{equation}
    P_r^{\,j}(e)=
    \frac{\bigl|\{v_i\in\mathcal{N}_r \mid e_i^{j}=e\}\bigr|}
    {\bigl|\mathcal{N}_r\bigr|},
\end{equation}
where $e_i^{j} \in \mathcal{E}$ denotes the cultural emotion label of image $v_i$ for culture $c_j$.
The dominant emotion of $q_r$ in $c_j$ is then the most probable one, $\bar{e}_r^{\,j}=\arg\max_{e\in\mathcal{E}}P_r^{\,j}(e)$.

\begin{table*}[t]
\centering
\footnotesize
\setlength{\tabcolsep}{3.0pt}
\renewcommand{\arraystretch}{1.01}
\caption{Accuracy of original models (Orig.), retrieval-augmented framework (+Know.), and SFT. The best result for each model within each culture is in \textbf{bold}. Subscripts denote absolute gains over Orig.}
\label{tab:improve-culture}
\begin{tabular}{ll>{\columncolor{gray!12}}ccc>{\columncolor{gray!12}}ccc>{\columncolor{gray!12}}ccc>{\columncolor{gray!12}}ccc}
\toprule
\multirow{2}{*}{Source}
& \multirow{2}{*}{Model}
& \multicolumn{3}{c}{English}
& \multicolumn{3}{c}{Chinese}
& \multicolumn{3}{c}{Arabic}
& \multicolumn{3}{c}{Overall} \\
\cmidrule(lr){3-5}
\cmidrule(lr){6-8}
\cmidrule(lr){9-11}
\cmidrule(lr){12-14}
& &
Orig. & +Know. & SFT &
Orig. & +Know. & SFT &
Orig. & +Know. & SFT &
Orig. & +Know. & SFT \\
\midrule
\multirow{11}{*}{Open}
& Qwen3-VL-8B
& 50.37 & 55.67 & \textbf{65.97}
& 41.38 & 54.34 & \textbf{75.99}
& 32.99 & 37.26 & \textbf{60.53}
& 41.58 & 49.09\gain{7.51} & \textbf{67.50}\gain{25.92} \\
& Aya-Vision-8B
& 54.34 & 58.32 & \textbf{64.95}
& 33.87 & 63.33 & \textbf{74.37}
& 29.16 & 31.81 & \textbf{58.62}
& 39.13 & 51.15\gain{12.02} & \textbf{65.98}\gain{26.85} \\
& InternVL3.5-8B
& 56.26 & 58.03 & \textbf{65.24}
& 54.34 & 61.71 & \textbf{73.78}
& 30.34 & 38.59 & \textbf{59.35}
& 46.98 & 52.77\gain{5.79} & \textbf{66.13}\gain{19.15} \\
& Qwen3.5-9B
& 55.23 & 58.47 & \textbf{68.34}
& 43.89 & 61.86 & \textbf{75.70}
& 31.22 & 41.68 & \textbf{61.12}
& 43.45 & 54.00\gain{10.55} & \textbf{68.38}\gain{24.93} \\
& Pixtral-12B
& 52.72 & 55.38 & \textbf{62.89}
& 17.08 & 55.23 & \textbf{75.11}
& 27.69 & 31.66 & \textbf{59.35}
& 32.50 & 47.42\gain{14.92} & \textbf{65.78}\gain{33.28} \\
& Gemma-3-12B-It
& 53.76 & 55.96 & \textbf{65.83}
& 36.23 & 53.61 & \textbf{76.88}
& 28.13 & 33.87 & \textbf{60.09}
& 39.37 & 47.82\gain{8.45} & \textbf{67.60}\gain{28.23} \\
& InternVL3.5-14B
& 58.32 & 58.91 & \textbf{66.42}
& 51.10 & 59.50 & \textbf{74.81}
& 36.52 & 40.21 & \textbf{58.62}
& 48.65 & 52.87\gain{4.22} & \textbf{66.62}\gain{17.97} \\
& Qwen3.5-27B
& 58.32 & 59.20 & \textbf{66.13}
& 45.36 & 63.62 & \textbf{76.58}
& 41.53 & 42.42 & \textbf{61.71}
& 48.40 & 55.08\gain{6.68} & \textbf{68.14}\gain{19.74} \\
& Qwen3.6-27B
& 57.58 & 58.76 & \textbf{67.16}
& 38.73 & 63.92 & \textbf{76.88}
& 38.00 & 44.18 & \textbf{61.56}
& 44.77 & 55.62\gain{10.85} & \textbf{68.53}\gain{23.76} \\
& Gemma-4-31B-It
& 59.35 & 59.94 & \textbf{65.54}
& 46.24 & 63.33 & \textbf{76.29}
& 38.44 & 41.24 & \textbf{59.65}
& 48.01 & 54.84\gain{6.83} & \textbf{67.16}\gain{19.15} \\
& Qwen3.6-35B-A3B
& 55.38 & 59.06 & \textbf{65.68}
& 43.00 & 63.18 & \textbf{75.26}
& 39.76 & 41.97 & \textbf{60.38}
& 46.05 & 54.74\gain{8.69} & \textbf{67.11}\gain{21.06} \\
\midrule
\multirow{5}{*}{Closed}
& Claude Sonnet 4.6
& 58.62 & \textbf{58.91} & --
& 45.21 & \textbf{58.47} & --
& 35.94 & \textbf{40.65} & --
& 46.59 & \textbf{52.68}\gain{6.09} & -- \\
& Claude Opus 4.8
& 59.20 & \textbf{60.61} & --
& 51.33 & \textbf{61.18} & --
& 33.04 & \textbf{38.81} & --
& 47.86 & \textbf{53.56}\gain{5.70} & -- \\
& Gemini 3 Flash
& 58.62 & \textbf{59.14} & --
& 51.25 & \textbf{66.08} & --
& 39.18 & \textbf{42.92} & --
& 49.68 & \textbf{56.05}\gain{6.37} & -- \\
& Gemini 3.5 Flash
& 59.20 & \textbf{60.53} & --
& 50.81 & \textbf{65.24} & --
& 39.62 & \textbf{43.89} & --
& 49.88 & \textbf{56.55}\gain{6.67} & -- \\
& GPT-5.5
& 58.62 & \textbf{60.74} & --
& 41.83 & \textbf{61.48} & --
& 31.96 & \textbf{37.81} & --
& 44.13 & \textbf{53.33}\gain{9.20} & -- \\
\bottomrule
\end{tabular}
\end{table*}

 3) \textit{MLLM-Based Rationale Generation.}
 This step produces cultural rationales for each concept's dominant emotion.
For each concept $q_r$ and culture $c_j$, we prompt Gemini~3.5 Flash with three inputs: the concept, target culture, and emotion distribution $\mathbf{P}_r^{\,j}=\{P_r^{\,j}(e)\}_{e\in\mathcal{E}}$. 
Given these inputs, the model identifies the dominant emotion and generates rationales explaining why this emotion is associated with the concept in the given culture,

\begin{equation}
    \mathcal{R}_r^{\,j}=MLLM\bigl(q_r,\,c_j,\,\mathbf{P}_r^{\,j}\bigr),
\end{equation}
where  $\mathcal{R}_r^{j}$ denotes the generated rationale set.

 4) \textit{Reverse Emotion Validation for Knowledge Filtering.}
Since MLLM-produced rationales are prone to hallucinations,  we filter the tuples to preserve trustworthy knowledge.
Our core validation strategy relies on reverse emotion reasoning: a faithful cultural rationale should independently recover the matched emotion it explains. To eliminate bias inherited from the rationale-generating model,  we adopt two  MLLM judges (GPT-5.5 and Claude Opus 4.8).  We supply each judge solely with the rationale set $\mathcal{R}_r^{j}$, and prompt them to infer the underlying emotion. 
A tuple is flagged if either judge predicts an emotion different from the dominant emotion  $\bar{e}_r^{\,j}$, and all flagged tuples are further reviewed by native-speaker annotators for final verification.
The remaining tuples constitute the cultural emotion knowledge base $\mathcal{K}$, which contains $8{,}439$ tuples and $28{,}146$ rationales, distributed evenly across the three cultures ($2{,}784$, $2{,}902$, and $2{,}753$ tuples with $9{,}538$, $9{,}626$, and $8{,}982$
rationales for English, Chinese, and Arabic, respectively).
The resulting candidate tuples are represented as $(q_r,c_j,\bar{e}_r^{j},\mathcal{R}_r^{j})$. 
Detailed statistics are provided in the supplementary material.

\subsubsection{Retrieval-augmented Inference.}
\label{sec:inference}
Built on the knowledge base, we design a training-free inference pipeline that proceeds in three steps: extracting the concepts of the test image, retrieving the matched tuples from the knowledge base, and predicting with the retrieved tuples as context.

 1) \textit{Visual concept extraction.}
Given a test image $\check{v}$ and target culture $c_j$, we prompt the frozen MLLM to extract salient concepts, obtaining a concept set $\check{\mathcal{P}}=\{\check{p}_1, \dots,\check{p}_n\}$, where $n$ denotes the number of extracted concepts.

\begin{table}[t]
\centering
\footnotesize
\setlength{\tabcolsep}{7.7pt}
\renewcommand{\arraystretch}{1}
\caption{Generalization to the unseen CEDAR benchmark: accuracy of Qwen3.6-27B with the two paradigms. The best result  is in \textbf{bold} and the second best is \underline{underlined}.}
\label{tab:cedar}
\begin{tabular}{lcccc}
\toprule
Model  & English & Chinese & Arabic & Overall \\
\midrule
\rowcolor{gray!12}
Qwen3.6-27B & \underline{44.50} & 13.75 & 27.75 & 28.66 \\
\ \ +Know. & 37.50 & \underline{23.25} & \underline{36.25} & \underline{32.33} \\
\ \ SFT & \textbf{49.75} & \textbf{39.25} & \textbf{38.50} & \textbf{42.50} \\
\bottomrule
\end{tabular}
\end{table}

  2) \textit{Exact-to-semantic knowledge retrieval.}
We then use the extracted concepts to query the knowledge base.
Since the MLLM may express a concept using different wording or granularity from the knowledge base, such as ``bamboo grove'' versus ``bamboo'', exact matching may miss useful tuples.
Therefore, we perform two-stage retrieval. Given the target culture $c_j$, for each concept $\check{p}_h$, we first search $\mathcal{K}_j$, the subset of knowledge tuples associated with $c_j$, for an exact concept match. If no exact match exists, we compute the semantic similarity between $\check{p}_h$ and each tuple concept using a sentence embedding model, and select the most similar tuple if its similarity exceeds a threshold $\tau$; otherwise, we discard $\check{p}_h$.
The retrieved tuples form $\mathcal{T}_{\check{v}}^{\,j}$, which serves as the cultural prior of the image $\check{v}$ under the culture $c_j$.

 3) \textit{Knowledge-conditioned emotion understanding.}
Finally, we convert  retrieved tuples $\mathcal{T}_{\check{v}}^{\,j}$ into a text sequence for prompt injection.  Conditioned on the image, target culture, and the available cultural knowledge, the frozen MLLM  predicts the emotion $\hat{e}$ and generates the explanation $\hat{s}$,
\begin{equation}
    (\hat{e},\,\hat{s})=MLLM\bigl(\check{v},\,c_j,\,\mathcal{T}_{\check{v}}^{\,j}\bigr).
\end{equation}
If no tuple is retrieved, the MLLM performs prediction without the injected cultural knowledge.

\subsection{Supervised Fine-tuning}
\label{sec:sft}
In this paradigm, instead of retrieving knowledge at inference time, we directly fine-tune open-source MLLMs on the ArtECulture training split, allowing the cultural emotion knowledge to be implicitly absorbed into model parameters. 
To leverage all available explanations, we construct a separate training instance for each explanation.
The input consists of the image $v_i$ and a prompt specifying the target culture $c_j$ and requesting an emotion label and explanation, while the target is the corresponding sequence $y=[e_i^j;s_{i,k}^{j}]$.
The model is fine-tuned with the standard autoregressive objective and directly outputs the emotion and corresponding explanation for the given image and target culture at inference.

\section{Experiments}
\label{sec:exp}


\subsection{Experimental Setup}
For emotion prediction, we report accuracy against culture-level labels. 
For explanation generation, we employ two MLLM judges, Claude Sonnet 5 (C5) and GLM-4.5V (G4V), to score each explanation on \textit{emotion alignment}, which measures how well the explanation supports the predicted emotion of the target culture, on a five-point scale. 
Implementation details are  in the supplementary material.

\begin{table}[t]
\centering
\footnotesize
\setlength{\tabcolsep}{3.5pt}
\renewcommand{\arraystretch}{1}
\caption{Ablation results on Qwen3.6-27B (Accuracy \%). }
\label{tab:ablation}
\begin{tabular}{lcccc}
\toprule
Setting & English & Chinese & Arabic & Overall \\
\midrule
\rowcolor{gray!12}
Orig. & 57.58 & 38.73 & 38.00 & 44.77 \\
+Know. & \textbf{58.76} & \textbf{63.92} & \textbf{44.18} & \textbf{55.62} \\
\midrule
+Internal-Know. & 55.67 & 47.57 & 33.28 & 45.51 \\
+Random-Know. & \underline{57.58} & \underline{55.08} & 35.49 & 49.39 \\
+Cross-Culture-Know. & 54.93 & 53.61 & \underline{41.68} & \underline{50.07} \\
\bottomrule
\end{tabular}
\end{table}

\begin{table}[t]
\centering
\footnotesize
\setlength{\tabcolsep}{16pt}
\renewcommand{\arraystretch}{1.0}
\caption{{Accuracy of Qwen3.6-27B under all combinations of prompt language (rows) and target culture (columns).}}
\label{tab:prompt-lang}
\begin{tabular}{lccc}
\toprule
Prompt & English & Chinese & Arabic \\
\midrule
English & \cellcolor{gray!12}57.58 & 49.63 & 37.85 \\
Chinese & 49.04 & \cellcolor{gray!12}38.73 & 30.04 \\
Arabic  & 48.46 & 36.38 & \cellcolor{gray!12}38.00 \\
\bottomrule
\end{tabular}
\end{table}

\begin{table}[t]
\centering
\caption{Explanation quality judged by Claude Sonnet 5 (C5) and GLM-4.5V (G4V) for Orig., +Know., and SFT. }
\label{tab:judge-claude}
\footnotesize
\setlength{\tabcolsep}{3pt}
\renewcommand{\arraystretch}{1}
\begin{tabular}{@{}l>{\columncolor{gray!12}}ccc>{\columncolor{gray!12}}ccc@{}}
\toprule
\multirow{2}{*}{Model}
& \multicolumn{3}{c}{C5 Align.}
& \multicolumn{3}{c}{G4V Align.} \\
\cmidrule(lr){2-4}\cmidrule(lr){5-7}
& Orig. & +Know. & SFT & Orig. & +Know. & SFT \\
\midrule
\multicolumn{7}{@{}l}{\textit{Open-source}} \\
Qwen3-VL-8B      & 3.26 & \textbf{3.35} & 2.43 & 3.37 & \textbf{3.56} & 2.96 \\
Aya-Vision-8B    & \textbf{3.22} & 3.10 & 2.46 & 3.27 & \textbf{3.49} & 2.99 \\
InternVL3.5-8B   & 3.00 & \textbf{3.02} & 2.36 & 3.35 & \textbf{3.41} & 2.89 \\
Qwen3.5-9B       & 3.34 & \textbf{3.36} & 2.49 & 3.46 & \textbf{3.68} & 3.01 \\
Pixtral-12B      & 2.83 & \textbf{3.01} & 2.38 & 3.13 & \textbf{3.35} & 2.91 \\
Gemma-3-12B-It   & 3.22 & \textbf{3.31} & 2.45 & 3.32 & \textbf{3.54} & 2.99 \\
InternVL3.5-14B  & 3.04 & \textbf{3.11} & 2.42 & 3.40 & \textbf{3.50} & 2.93 \\
Qwen3.5-27B      & \textbf{3.51} & 3.46 & 2.52 & 3.58 & \textbf{3.74} & 3.01 \\
Qwen3.6-27B      & 3.42 & \textbf{3.54} & 2.50 & 3.50 & \textbf{3.77} & 3.02 \\
Gemma-4-31B-It   & 3.47 & \textbf{3.55} & 2.51 & 3.56 & \textbf{3.71} & 3.00 \\
Qwen3.6-35B-A3B  & 3.50 & \textbf{3.60} & 2.53 & 3.55 & \textbf{3.76} & 3.00 \\
\midrule
\multicolumn{7}{@{}l}{\textit{Closed-source}} \\
Claude Sonnet 4.6 & 3.77 & \textbf{3.78} & -- & 3.65 & \textbf{3.83} & -- \\
Claude Opus 4.8   & 3.56 & \textbf{3.69} & -- & 3.45 & \textbf{3.73} & -- \\
Gemini 3 Flash    & \textbf{3.62} & 3.56 & -- & 3.58 & \textbf{3.72} & -- \\
Gemini 3.5 Flash  & \textbf{3.74} & 3.66 & -- & 3.71 & \textbf{3.83} & -- \\
GPT-5.5           & 3.42 & \textbf{3.55} & -- & 3.42 & \textbf{3.67} & -- \\
\bottomrule
\end{tabular}
\end{table}

\begin{table}[t]
\centering
\caption{Pairwise human evaluation of emotion alignment on $100$ randomly sampled test images for Qwen3.6-27B.}
\label{tab:human-eval}
\footnotesize
\setlength{\tabcolsep}{10pt}
\renewcommand{\arraystretch}{1.0}
\begin{tabular}{@{}llccc@{}}
\toprule
Comparison & Culture & Win & Tie & Loss \\
\midrule
\multirow{3}{*}{+Know.\ vs.\ Orig.}
 & English & 15.67 & 65.33 & 19.00 \\
 & Chinese & 41.67 & 32.33 & 26.00 \\
 & Arabic  & 38.33 & 31.00 & 30.67 \\
\midrule
\multirow{3}{*}{SFT vs.\ Orig.}
 & English & 16.33 & 25.00 & 58.67 \\
 & Chinese & 12.67 & 11.33 & 76.00 \\
 & Arabic  & 26.00 & 6.00  & 68.00 \\
\bottomrule
\end{tabular}
\end{table}

\subsection{Culture-Conditioned Emotion Prediction}
\label{sec:exp-prediction}



\paragraph{Effect of the two paradigms.}
Table~\ref{tab:improve-culture} compares the two paradigms  for the $16$ MLLMs in per-culture and overall accuracy. For reference, we also incorporate the zero-shot performance, denoted as {Orig.}.
1) Based on the zero-shot performance, we noted that
all MLLMs struggle with the task: even the best one, Gemini 3.5 Flash, attains an overall accuracy of only $49.88\%$, and most fall below $48\%$. Moreover, the performance is highly imbalanced across cultures.
Every model achieves its highest accuracy on English, while accuracy on Arabic is consistently the lowest for nearly all models, at best merely $41.53\%$, suggesting that the affective conventions absorbed during pre-training mainly come from English-dominated corpora. The finer breakdown performance of each culture on Western and non-Western images reveals an asymmetric pattern, as detailed in the supplementary material.
2) +Know. yields consistent and substantial accuracy gains across all models without extra training, 
demonstrating that the knowledge base compensates for missing parametric cultural knowledge.
Notably, the gains of +Know.\ are highly uneven across cultures, concentrating on Chinese, moderate on Arabic, and marginal on English.
This pattern is consistent with  the English-centric priors: external knowledge is  redundant for English but fills a substantial gap for the other cultures. {A possible reason why Chinese benefits more than Arabic is that the rationales are written by an MLLM whose cultural coverage  is likely richer for Chinese than for Arabic.} 
3)  SFT brings much larger gains, showing  in-domain training is currently the most effective way to acquire cultural knowledge for emotion prediction.

\paragraph{Generalization of the two paradigms.}
For generalization evaluation, we evaluate both paradigms using Qwen3.6-27B on the multimodal subset of CEDAR~\cite{DBLP:conf/acl/DaiSGLJWLGH26}, an unseen benchmark whose image styles and question formats differ from ArtECulture. As shown in Table~\ref{tab:cedar}, both paradigms improve the overall accuracy.
Retrieval augmentation raises  accuracy on Chinese and Arabic but reduces it on English, likely because the model is already better aligned with English emotional conventions, leaving limited room for additional gains from retrieved knowledge. In such cases, imperfectly matched retrieved knowledge may occasionally introduce noise. SFT improves all three cultures and achieves the best overall accuracy, indicating that the fine-tuned model learns transferable cultural emotion knowledge rather than dataset-specific patterns.

\begin{figure*}[!t]
    \centering
    \includegraphics[scale=0.3]{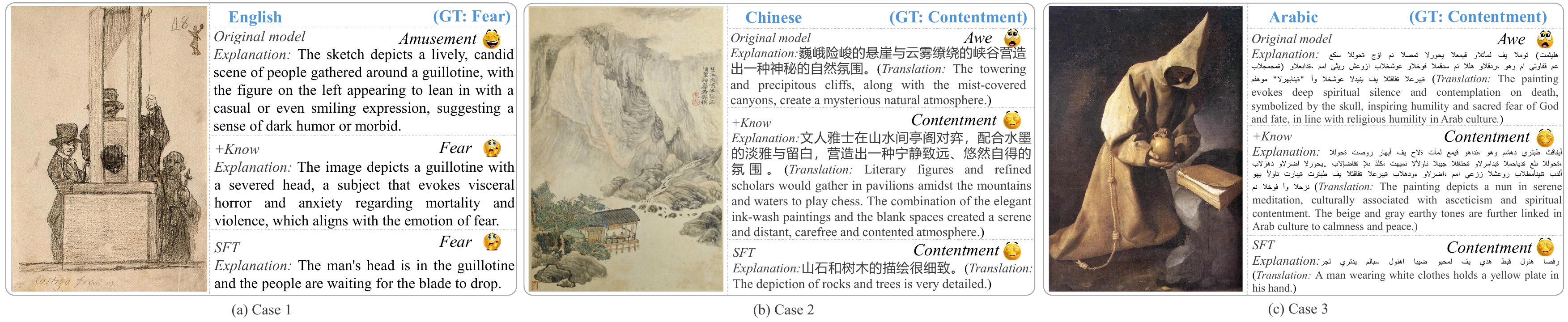}
\caption{Case study of Orig., +Know., and SFT with Qwen3.6-27B on three test images from different cultures. Each case includes the ground-truth emotion (GT), predictions, and explanations. Arabic and Chinese outputs are translated into English.}
    \label{fig:case}
\end{figure*}

\paragraph{Ablation study.}
To validate our proposed retrieval-augmented paradigm, we compare +Know.\ with three variants on Qwen3.6-27B in Table~\ref{tab:ablation}: 1) \textbf{+Internal-Know.}, which relies solely on its internal parametric knowledge with chain-of-thought (CoT) prompting for emotion reasoning without the external knowledge base; 2) \textbf{+Random-Know.}, which replaces the retrieved tuples with randomly sampled ones; and 3) \textbf{+Cross-Culture-Know.}, which injects tuples retrieved for a mismatched culture. 
We draw three observations.
1) +Internal-Know.\ improves the overall accuracy by only $0.74$ points and even reduces the Arabic accuracy, suggesting that chain-of-thought prompting can elicit useful internal reasoning but remains insufficient to provide culture-specific emotion knowledge, particularly for underrepresented cultures. 
{2) +Random-Know.\ reaches $49.39\%$, as injecting any external knowledge biases the predictions toward the benchmark's emotion prior, and +Cross-Culture-Know.\ performs slightly better at $50.07\%$, since concepts often share the same dominant emotion across cultures.}
3) +Know.\ outperforms +Cross-Culture-Know.\ by $5.55$ points overall and $10.31$ points on Chinese, showing that a substantial part of the gain comes from the culture-specific knowledge itself and cannot be attributed to the injection mechanism alone.

\paragraph{Effect of prompt language.}
Since the main experiments use prompts in the target culture's language, cross-cultural performance gaps may partly come from the prompt language rather than from differences in cultural emotion knowledge.  
To examine this, we evaluate Qwen3.6-27B under all nine combinations of prompt language and target culture in Table~\ref{tab:prompt-lang}. 
Accuracy is consistently highest under the English culture condition regardless of the prompt language, while changing the prompt language produces smaller and less uniform effects.
These patterns are consistent with an English-centric language bias and a Western-centric cultural knowledge bias in current MLLMs, indicating that the performance gap across cultures cannot be explained solely by prompt language.

\subsection{Culture-Conditioned Explanation Generation}
\label{sec:exp-explanation}
We evaluate explanation generation with the two MLLM judges on all $16$ MLLMs, and a human study on a subset. 

\paragraph{Automatic evaluation.}
Table~\ref{tab:judge-claude} reports the  evaluation results. We highlight two findings. 
1) +Know.\ improves emotion alignment for all 16 models under G4V and most models under C5, showing that retrieval-augmented knowledge enhances both prediction accuracy and explanation quality. This is because retrieved cultural knowledge provides emotion-related associations that guide explanations toward the predicted emotion.
2) In contrast to its superiority in emotion prediction, SFT yields the lowest alignment scores among the three settings. 
We investigate this degradation and find that the average explanation length of open-source models drops from $30.50$ to $15.12$  words after SFT. 
This is likely caused by the multiple-reference training setup: each image-culture pair is associated with multiple explanations, while each explanation is used as an independent target during SFT. 
Under token-level likelihood optimization, the model tends to converge to short, generic explanations that are compatible with multiple references, sacrificing image- and culture-specific details.
We further measure the agreement between  two judges. Ranking  the $16$ models within each setting, the Spearman correlation between the two judges ranges from $0.860$ to $0.905$ (all $p<0.001$), confirming the reliability of the automatic evaluation.

\paragraph{Human evaluation.}
We further conduct a pairwise human study using Qwen3.6-27B. For $100$ randomly sampled test images, native speakers compare two anonymized explanations  in random order, and each pair is rated by three annotators per culture. 
As shown in Table~\ref{tab:human-eval}, the human judgments follow the same overall trend as the automatic evaluation in Table~\ref{tab:judge-claude}: +Know.\ outperforms Orig.\ while SFT falls behind both. 
Specifically, +Know.\ receives more wins than losses on Chinese ($41.67\%$ against $26.00\%$) and Arabic ($38.33\%$ against $30.67\%$). 
On English, +Know.\ and Orig.\ are comparable, and most comparisons end in a tie ($65.33\%$). 
This is consistent with the accuracy results in Table~\ref{tab:improve-culture}: the model already captures English emotional conventions relatively well, so external knowledge brings limited  gains.

\paragraph{Case study.}
Figure~\ref{fig:case} compares Orig., +Know., and SFT with Qwen3.6-27B on three test images, one per culture.
Orig.\ mainly relies on visible content and overlooks the underlying cultural meanings, interpreting the guillotine scene as dark humor, the misty cliffs as mysterious grandeur, and the praying figure as fear of death.
In contrast, +Know.\ incorporates retrieved cultural knowledge and yields more culture-aware interpretations, recognizing the guillotine as a symbol of violence and mortality, the ink-wash scene as reflecting the traditional Chinese appreciation of peaceful living, and the monk's meditation with earthy tones as spiritual calmness in Arab culture.
SFT predicts the correct emotion in all three cases, but its explanations become short and generic (e.g., ``the depiction of rocks and trees is very detailed''), lacking cultural reasoning for the emotion. This reveals the trade-off: SFT improves prediction accuracy at the cost of generating less culturally grounded explanations.

\section{Conclusion}
We study culture-conditioned visual emotion understanding, which requires predicting the emotion an image evokes in a given culture and explaining the underlying rationale.
To support this task, we construct ArtECulture, the first benchmark that provides culture-level emotion labels for three cultures with a regionally balanced visual distribution.
We propose a training-free retrieval-augmented framework built on a concept-level cultural emotion knowledge base and compare it with supervised fine-tuning, revealing a complementary trade-off between prediction accuracy and explanation quality.
Future work includes extending the benchmark to more cultures, adopting distribution-level labels that preserve within-group disagreement, and combining the strengths of both paradigms.

\clearpage

\bibliography{reference}
\clearpage

\end{document}